\ifdefined\pdfminorversion\pdfminorversion=7\fi
\PassOptionsToPackage{hyphens}{url}
\documentclass[11pt, letterpaper, logo, onecolumn, copyright]{hunyuan_report}
\usepackage[numbers,sort&compress]{natbib}

\usepackage{amsmath,amsfonts,bm}

\def\eqref#1{equation~\ref{#1}}
\def\1{\bm{1}}

\DeclareMathAlphabet{\mathsfit}{\encodingdefault}{\sfdefault}{m}{sl}
\SetMathAlphabet{\mathsfit}{bold}{\encodingdefault}{\sfdefault}{bx}{n}

\usepackage{float}
\usepackage{booktabs}
\usepackage{dcolumn}
\usepackage{multirow}
\usepackage{graphicx}
\usepackage{float}
\usepackage{placeins}
\usepackage{xcolor}
\usepackage{colortbl}
\usepackage{titletoc}
\usepackage{hyperref}
\hypersetup{hidelinks}
\usepackage{url}
\usepackage{listings}
\usepackage[breakable,skins]{tcolorbox}

\lstdefinestyle{prompt}{
  basicstyle=\ttfamily\footnotesize,
  breaklines=true,
  columns=fullflexible,
  keepspaces=true,
  frame=single,
  rulecolor=\color{black!25},
  xleftmargin=0.35em,
  xrightmargin=0.35em,
  aboveskip=0.55em,
  belowskip=0.55em,
  showstringspaces=false
}

\definecolor{realhopblue}{HTML}{1F4E79}
\definecolor{rankup}{HTML}{2F6B45}
\definecolor{rankdown}{HTML}{9A4A45}
\definecolor{rankneutral}{HTML}{6E6E6E}
\definecolor{rankgold}{HTML}{A87912}
\definecolor{ranksilver}{HTML}{6F7780}
\definecolor{rankbronze}{HTML}{95613C}
\definecolor{originalbg}{HTML}{F3F3F1}
\definecolor{realhopbg}{HTML}{EDF4F8}
\definecolor{realhopsummarybg}{HTML}{E1EDF5}

\newcommand{\rankfirst}{\textcolor{rankgold}{\textbf{1}}}
\newcommand{\ranksecond}{\textcolor{ranksilver}{\textbf{2}}}
\newcommand{\rankthird}{\textcolor{rankbronze}{\textbf{3}}}

\newcommand{\realhop}{\textsc{RealHop}}
\newcommand{\realhopmusique}{\textsc{RealHop-MuSiQue}}
\newcommand{\realhopframes}{\textsc{RealHop-FRAMES}}
\newcommand{\realhoplongbench}{\textsc{RealHop-LongBench}}

\title{\realhop: Rethinking Multi-Hop Reasoning Evaluation via Behavioral Auditing}
\runningtitle{\realhop: Behavioral Auditing for Multi-Hop Reasoning}
\author{
  Jiawen Tao\textsuperscript{\rm 1,2}\textsuperscript{*},
  Xiaokun Yuan\textsuperscript{\rm 1,2}\textsuperscript{*},
  Yaoming Li\textsuperscript{\rm 2}\\
  Chenxu Liu\textsuperscript{\rm 1},
  Mengzhou Wu\textsuperscript{\rm 1},
  Tong Yang\textsuperscript{\rm 2}\textsuperscript{\textdagger},
  Maxm Pan\textsuperscript{\rm 1}\textsuperscript{\textdagger}\\
  \vspace{0.3cm}
  \normalsize
  \textsuperscript{1}Hunyuan Team, Tencent\\
  \textsuperscript{2}Peking University\\
  {\fontfamily{lmtt}\selectfont
  \{jadentao,xiaokunyuan\}@tencent.com}
}
\date{}

\begin{abstract}
Complex questions often require multi-hop reasoning that connects facts
distributed across sources or distant regions of a long context through
intermediate steps.
Benchmarks commonly evaluate this ability with questions built around
predefined reasoning chains, treating a correct answer as evidence that the
intended composition was used. Yet answer correctness alone leaves open whether success depends on the evidence associated with each intended step: models may instead rely on memorized
associations, shorter paths, or partial evidence. We examine this dependence using the Behavioral Necessity Rate (BNR), which measures how often targeted evidence
removal prevents answer recovery on initially correct instances. Across five existing benchmarks, panel-mean BNR is
only 16.6--48.9\%, exposing a substantial gap between annotated structure and
observed dependence. Guided by this diagnosis, we introduce \realhop{}, a
diagnose--construct--verify framework that rebinds entities, factorizes selected
relations, adds complete competing paths, and places evidence at traceable
locations. Structural and semantic checks precede freezing; behavioral
interventions follow. On 790 paired MuSiQue questions, \realhop{} raises
panel-mean BNR from 27.4\% to 94.4\% while retaining high Full accuracy. It
also yields high BNR on \realhopframes{} and \realhoplongbench{}. On 216
long-context questions, the matched multiple-choice spread
across 16 models grows from 13.9 to 59.2 points and persists under repeated
open-ended evaluation. Together, these results show that a conceptually
coherent chain and a correct final answer do not by themselves establish
multi-hop reasoning. Verifying that success depends on every intended hop is
therefore as fundamental to multi-hop evaluation as measuring answer accuracy
itself.
\end{abstract}

\begin{document}
\raggedbottom

\maketitle
\begingroup
\renewcommand{\thefootnote}{*}
\footnotetext{Equal contribution.}
\renewcommand{\thefootnote}{\textdagger}
\footnotetext{Corresponding authors.}
\endgroup
\setcounter{footnote}{0}

\section{Introduction}
\label{sec:introduction}

Multi-hop reasoning is essential for meeting complex information needs: it
requires models to connect facts distributed across sources or distant regions
of a long context and use intermediate reasoning steps to answer questions
that no single piece of information can resolve
\citep{yang2018hotpotqa,ho2020twowiki,trivedi2022musique,bai2024longbench2}.
To evaluate this capability, researchers construct questions with predefined
reasoning chains and supporting evidence, such that information obtained at
one step supports subsequent inference and ultimately leads to the answer
\citep{yang2018hotpotqa,ho2020twowiki,trivedi2022musique}.  Answer accuracy is
therefore commonly interpreted as evidence that the model carried out the
intended composition.  This interpretation, however, raises two questions:
\textbf{Does model success require the evidence associated with each intended
step in the predefined chain?  And to what extent can success on multi-hop QA
benchmarks be taken as evidence of multi-hop reasoning ability?}

\begin{figure}[!t]
  \centering
  \includegraphics[width=\linewidth]{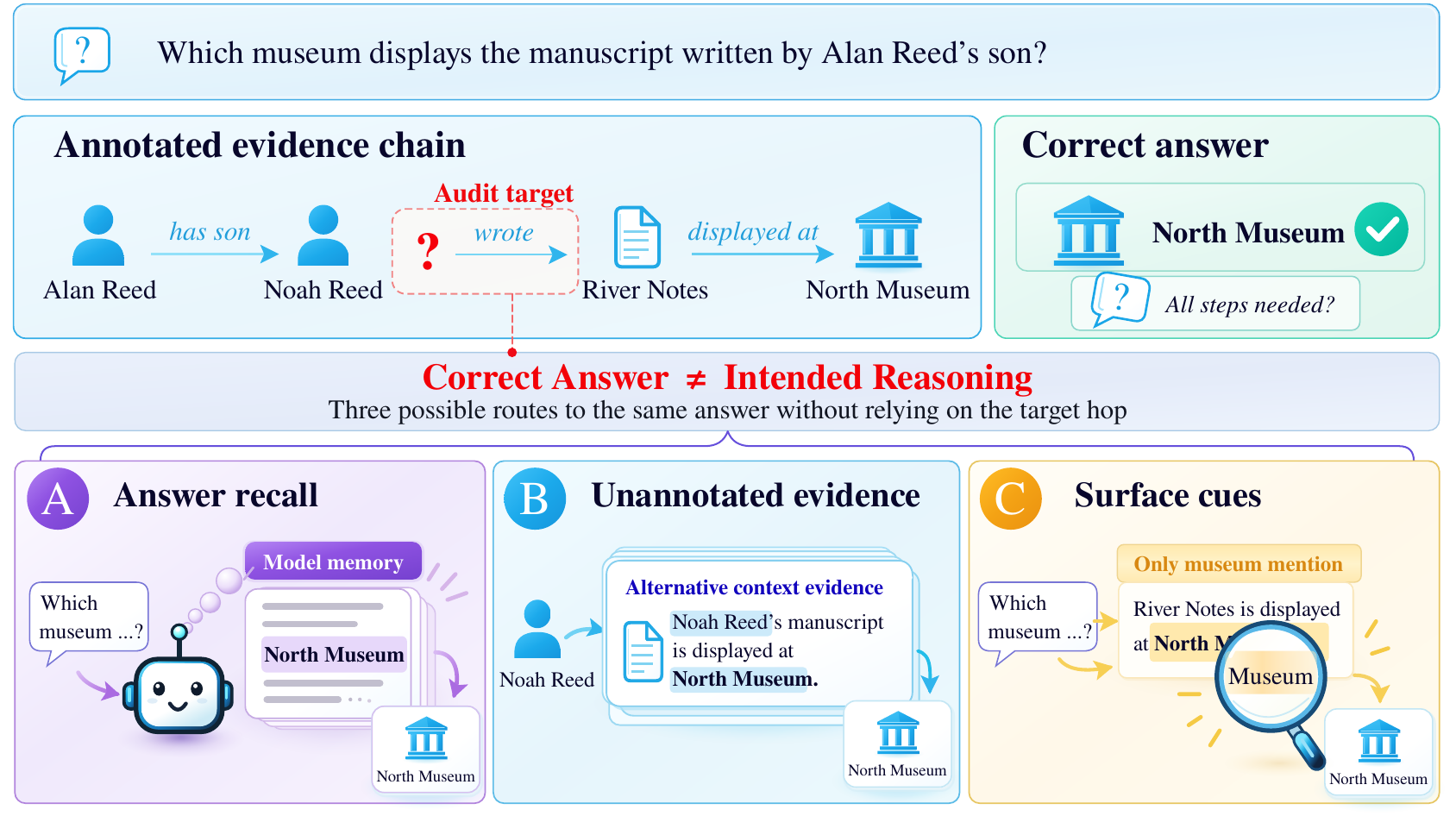}
  \caption{Three bypasses of an annotated gold chain: answer recall,
  unannotated evidence, and surface cues.  Each can preserve the answer after
  evidence for a gold hop is removed.}
  \label{fig:shortcut-teaser}
\end{figure}

These questions matter because models can bypass the predefined chain in
several ways: they may recall the answer, follow an unannotated shorter path,
or exploit a surface cue
\citep{min2019compositional,jiang2019avoiding}
(Figure~\ref{fig:shortcut-teaser}).  Existing construction techniques reduce
such shortcuts through connected composition, false-chain perturbations, and
plausible distractors
\citep{trivedi2022musique,serbanescu2025falsecotqa,bhuiya2024seemingly}, but
neither a coherent path nor lower accuracy establishes dependence: noise can
make a benchmark harder without making it more diagnostic.  A more direct way
to test evidence necessity is to intervene on the claimed supports and observe
whether model success changes.

We operationalize this idea through behavioral auditing based on targeted
evidence deletion.  Among
questions answered correctly under Full context, the \emph{Behavioral
Necessity Rate} (BNR) measures how often removing claimed evidence prevents
answer recovery; a matched-background Control separates evidence-specific
failure from generic deletion sensitivity.  Across MuSiQue, HotpotQA, and
2WikiMultiHopQA, panel-mean BNR is only 16.6--27.4\%.  Even targeted
adversarial constructions designed to make shortcut solving harder yield
panel-mean BNRs of only 43.5--48.9\%
\citep{serbanescu2025falsecotqa,bhuiya2024seemingly}.  Together, these results
show that neither explicit hop annotations nor targeted distractors ensure
that successful prediction depends on the intended evidence.

To address this gap, we propose \realhop{}, a diagnose--construct--verify
framework for building and testing multi-hop questions designed to require
the evidence for every reasoning step.
It first checks whether models can bypass annotated evidence in existing
benchmarks. It then replaces entities consistently, splits selected relations
into premises that work together, and adds complete paths to alternative
answers that fail a required relation. Each inserted fact is recorded at a
known text location. Structural and semantic checks determine which items
are retained before the dataset is frozen; subsequent evidence-deletion
tests measure whether models can answer with and without the intended
support. This separates benchmark construction from behavioral evaluation.

We evaluate this framework in three multi-hop settings.  On 790 paired MuSiQue
questions, \realhopmusique{} raises panel-mean BNR from 27.4\% to 94.4\%,
while Full accuracy remains 68.5--83.5\%.  On source-linked
FRAMES~\citep{krishna2025frames} questions without native hop annotations,
\realhopframes{} reaches 88.9--89.9\% BNR.  We then scale the construction to 216 long
context questions from LongBench v2~\citep{bai2024longbench2}.  Under a matched
multiple-choice protocol, \realhoplongbench{} expands the 16-model accuracy
spread from 13.9 to 59.2 points.  This wider separation persists in repeated
open-ended evaluation, while targeted deletion yields 96.5--98.1\% BNR across
three audited models.

\textbf{Our contributions are threefold:}
\begingroup
  \setlength{\leftmargini}{1.15em}
  \setlength{\labelsep}{0.4em}
  \setlength{\topsep}{0.4em}
  \setlength{\itemsep}{0.5em}
  \setlength{\parsep}{0pt}
  \setlength{\parskip}{0pt}
\begin{itemize}
  \item \textbf{We audit existing multi-hop benchmarks and reveal the gap
  between annotated reasoning structure and observed evidence dependence.}
  Annotated chains specify the intended reasoning, yet models
  can still answer correctly after evidence for a step is removed. We
  introduce BNR to quantify this gap across three standard benchmarks and
  two adversarial constructions.

  \item \textbf{We introduce \realhop{}, a diagnose--construct--verify framework
  for rebuilding benchmarks to make each reasoning step necessary for
  answering.} It combines entity rebinding, evidence rewriting, and complete
  competing paths, with structural and semantic checks before dataset
  freezing and behavioral evaluation afterwards.

  \item \textbf{We construct and evaluate three benchmark variants covering
  standard and long-context multi-hop question answering.} Built from
  MuSiQue, FRAMES, and LongBench v2, these variants are evaluated through
  targeted evidence deletion, component ablations, and repeated long-context
  trials across 16 models.
\end{itemize}
\endgroup

\section{Related Work}
\label{sec:related-work}

\paragraph{Standard multi-hop QA and shortcut-resistant evaluation.}
Multi-hop QA datasets annotate supporting chains so that answer correctness
can be read as evidence of connected reasoning
\citep{welbl2018constructing,yang2018hotpotqa,talmor2018web,ho2020twowiki,trivedi2022musique}.
Analyses of HotpotQA showed that compositional questions still admit
single-hop solutions, sentence-level cues, and word-matching routes,
including adversarial documents that preserve the original
answer~\citep{min2019compositional,chen2019understanding,jiang2019avoiding}.
Models can also know the individual hops yet fail to compose
them~\citep{press2023compositionality}.

Construction work then targeted those artifacts: MuSiQue composes connected
single-hop questions and filters disconnected
reasoning~\citep{trivedi2022musique}; StrategyQA and MoreHopQA add implicit or
generative hops~\citep{geva2021strategyqa,schnitzler2024morehopqa};
seemingly plausible distractors insert complete but incorrect multi-hop
chains~\citep{bhuiya2024seemingly}; and
FalseCoTQA and CRiT-QA inject knowledge-grounded or hop-anchored counterfactual
chains~\citep{serbanescu2025falsecotqa,yun2026critqa}.
Adversarial distractor methods change the evidence a model sees and typically
report the resulting accuracy drop to evaluate robustness.
Such methods can make shortcut solving harder, but an accuracy drop under
stronger distractors does not establish that successful predictions depend on
every intended hop.

\paragraph{Long-context multi-hop reasoning.}
Long contexts extend the multi-hop challenge: models must locate facts
dispersed across lengthy or multi-document inputs and then combine them along
a reasoning chain.  LongBench and LongBench v2 include multi-document and
reasoning-intensive tasks~\citep{bai2024longbench,bai2024longbench2}, while
Loong places multi-document questions in extended
carriers~\citep{wang2024loong}.  FRAMES provides source-linked questions that
require combining information across sources, but does not annotate their hop
chains~\citep{krishna2025frames}.  Controlled evaluations make the same
challenge explicit: RULER includes multi-hop tracing, and BABILong embeds
reasoning facts in long background
\citep{hsieh2024ruler,kuratov2024babilong}.  Synthetic long-context training
similarly hides multi-hop questions behind key chains or trajectory-derived
distractors
\citep{wang2025loongrl,lin2026longtracerl}.
These settings test whether models can find and combine information at scale,
but neither context length nor the presence of a chain establishes that
successful answers depend on every intended hop.  Long-context evaluation
therefore scales the multi-hop problem without resolving evidence necessity.

\paragraph{Evidence necessity through intervention.}
Sufficiency asks whether selected evidence is enough; necessity asks whether
a correct prediction survives its removal.
Deletion tests from interpretability make evidence dependence observable.
Extractive rationales treat a sparse subset of the input as sufficient for
prediction~\citep{lei2016rationalizing}; later work asks whether those
selections are also
comprehensive~\citep{yu2019rethinking,deyoung2020eraser}.
Attention weights and chain-of-thought traces remain unreliable proxies for
this dependence
\citep{jain2019attention,wiegreffe2019attention,jacovi2020towards,turpin2023unfaithful,lanham2023measuring}.
Within multi-hop QA, DiRe combines predictions on complementary support
subsets to measure disconnected reasoning~\citep{trivedi2020dire}.
\realhop{} directly intervenes on individual hops to measure whether a
successful prediction depends on each claimed support, while a matched
background deletion separates evidence-specific failure from generic deletion
sensitivity.  Unlike prior audits, this diagnosis feeds back into construction:
complete competing paths make intended relations consequential, and structural,
semantic, and behavioral checks verify the result.  The resulting
diagnose--construct--verify framework applies the same criterion of evidence
necessity across standard and long-context multi-hop settings.

\section{Problem Formulation and Benchmark Audit}
\label{sec:problem-and-audit}

\textbf{A model can answer correctly without relying on a benchmark's
annotated evidence structure.}  We formalize this gap and audit whether models
depend on each claimed evidence target.

\subsection{The Conceptual--Behavioral Gap}
\label{sec:necessity-gap}
An annotated evidence graph claims that its steps form a coherent and
sufficient structure supporting the answer, but does not establish that a
model must use every step.  Formally, an instance is
$x=(q,C,a^\star,G)$. Let $H$ index the intervention targets associated with
the claimed steps or support groups in $G$, and let $W_h\subseteq C$ denote
the evidence assigned to target $h\in H$. We intervene on $h$ by evaluating
the unchanged question on $C\setminus W_h$. \emph{Conceptual necessity} holds
when $C\setminus W_h$ contains no sufficient path to $a^\star$;
\emph{behavioral necessity} holds for a model when it answers correctly on
$C$ but not on $C\setminus W_h$. The former does not guarantee the latter:
a model may bypass a coherent annotated chain through shortcuts, answer cues,
unannotated evidence, or parametric knowledge.

\subsection{Audit Protocol and Metrics}
\label{sec:audit-metrics}
The \emph{Behavioral Necessity Rate} (BNR) compares \textsc{Full} with
\textsc{Drop-one}, which removes one declared evidence target at a time. For
a fixed model, we restrict attention to items with complete Drop evaluations,
sample item $i$ uniformly and target $h$ uniformly from $H_i$, and let $F_i$
indicate Full correctness, $D_{ih}$ Drop correctness, and
$\mathcal I^+=\{i:F_i=1\}$:
\begin{equation}
 \begin{aligned}
 \mathrm{BNR}
 &= \mathbb{E}[1-D_{ih}\mid F_i=1] \\
 &= \frac{1}{|\mathcal I^+|}
    \sum_{i\in\mathcal I^+}\frac{1}{|H_i|}
    \sum_{h\in H_i}(1-D_{ih}).
 \end{aligned}
 \label{eq:bnr}
\end{equation}
Thus, BNR is the item-macro proportion of testable targets whose removal
prevents recovery of the gold answer, without giving longer chains greater
weight. Because BNR is model- and protocol-specific, we report it alongside
Full accuracy. Length-matched background \textsc{Control} provides a separate
specificity check and is not an input to BNR.

\subsection{Audit of Existing Multi-Hop Benchmarks}
\label{sec:existing-benchmark-audit}
\paragraph{Multi-hop annotation substantially overstates observed dependence.}
We audit 790 source MuSiQue, 300 HotpotQA,
300 2WikiMultiHopQA questions, a stratified 300-item
FalseCoTQA--MuSiQue sample, and a 300-item Plausible Distractors split of
HotpotQA with DeepSeek V4 Flash, Gemini 3.6 Flash, and Qwen 3.7 Max.
Sampling and deletion units appear in
Appendix~\ref{app:external-multihop-per-model}.

\begin{table}[!t]
  \centering
  \small
  \caption{Panel-mean Full, Drop-one, Control, and BNR on existing multi-hop
  benchmarks (percent). Means give equal weight to DeepSeek V4 Flash,
  Gemini 3.6 Flash, and Qwen 3.7 Max. Drop-one and Control are accuracies
  over all audited questions, whereas BNR conditions on Full-correct ones.}
  \label{tab:external-multihop-summary}
  \setlength{\tabcolsep}{5pt}
  \renewcommand{\arraystretch}{1.12}
  \begin{tabular}{@{}lcccc@{}}
    \toprule
    Benchmark & Full & Drop-one & Control & BNR \\
    \midrule
    MuSiQue~\citep{trivedi2022musique} & 87.4 & 65.4 & 87.3 & 27.4 \\
    HotpotQA~\citep{yang2018hotpotqa} & 81.2 & 69.7 & 81.5 & 16.6 \\
    2WikiMultiHopQA~\citep{ho2020twowiki} & 88.6 & 74.5 & 89.1 & 17.0 \\
    FalseCoTQA~\citep{serbanescu2025falsecotqa} & 63.8 & 36.3 & 59.9 & 43.5 \\
    Plausible Distractors~\citep{bhuiya2024seemingly} & 87.8 & 44.8 & 86.4 & 48.9 \\
    \bottomrule
  \end{tabular}
\end{table}

Table~\ref{tab:external-multihop-summary} exposes a consistent gap. Control
remains near Full and well above Drop, supporting evidence-specific
sensitivity rather than an effect of deletion volume alone. Yet BNR is only
16.6--27.4\% on the three standard benchmarks. FalseCoTQA and Plausible
Distractors raise it to 43.5\% and 48.9\%, respectively, but correct answers
still survive most gold-hop deletions. These descriptive, unpaired results
remain far below the 93.9--95.0\% achieved by \realhopmusique{}
(Section~\ref{sec:paired-necessity-results}).

These findings motivate constructions that make every intended binding
consequential; Section~\ref{sec:method} describes how \realhop{} pursues this
goal in both standard and long-context multi-hop QA.

\section{Method}
\label{sec:method}

\begin{figure}[H]
  \centering
  \includegraphics[width=\linewidth]{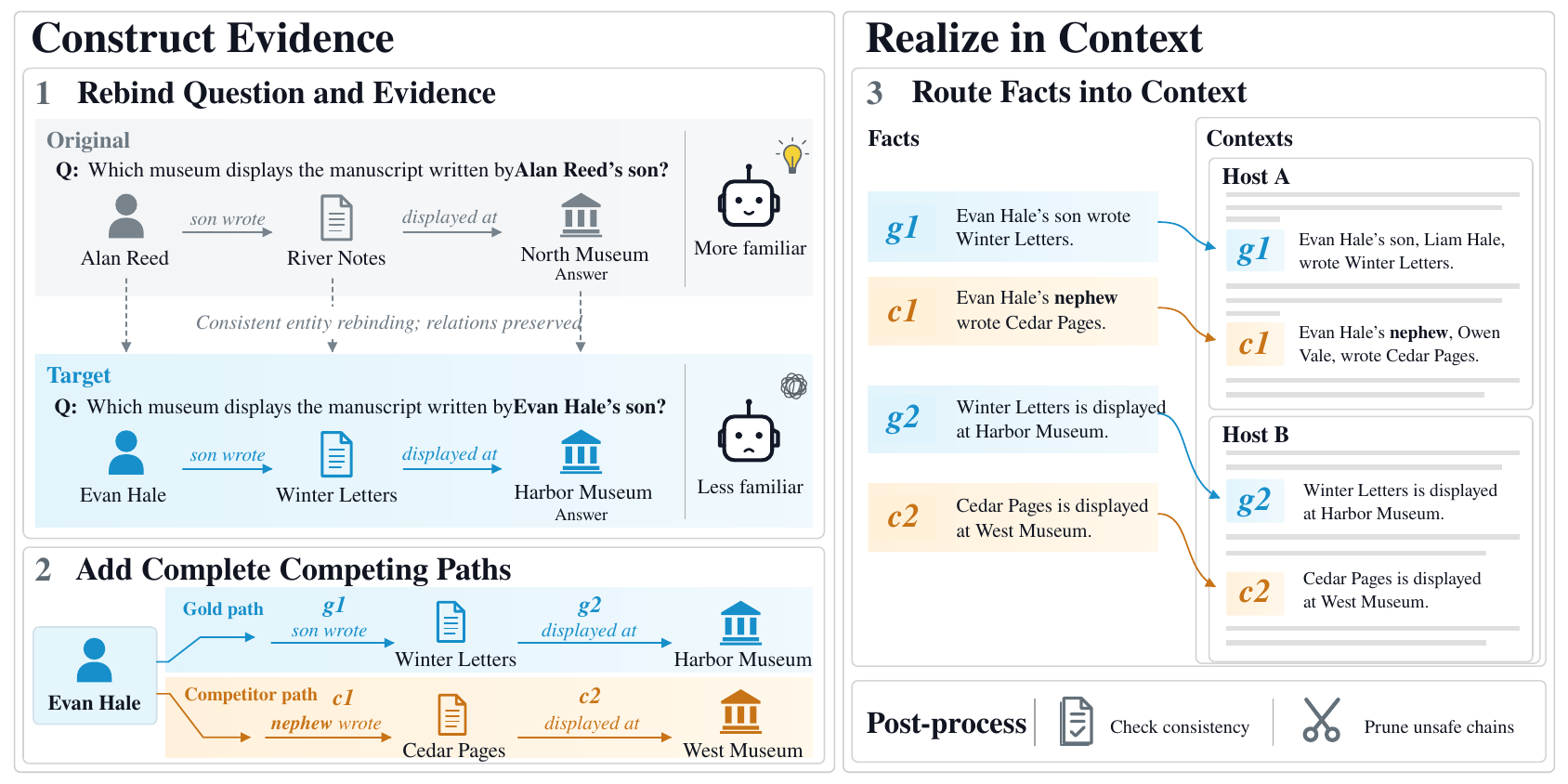}
  \caption{\textbf{Overview of the two-stage construct phase of \realhop{}.}
  The framework constructs correct and competing evidence paths, then realizes
  and validates them in context. Gray marks the original source graph, blue the
  correct path, and orange the competing path; matching labels link planned
  facts to their contextual realizations.}
  \label{fig:realhop-construction}
\end{figure}

This section details the construct phase of the \realhop{} framework, which
proceeds in two stages. Given a multi-hop question $q$, context $C$, answer
$a^\star$, and dependency graph $G$, it first \textbf{constructs evidence},
then \textbf{realizes it in context} (Figure~\ref{fig:realhop-construction}).
Models generate facts and text; separate checks determine sample acceptance.

\subsection{Construct Evidence}
\label{sec:compiler-interface}
\paragraph{Rebind question and evidence.}
We replace entities in the question, planned facts, and answer
with \emph{virtual counterparts}, preserving relations, entity roles,
answer type, and reasoning structure. In the figure, Alan Reed, River Notes,
and North Museum become Evan Hale, Winter Letters, and Harbor Museum.
This aims to prevent answer recall from prior knowledge and make models
reason from the provided context. The question does not reveal intermediate
answers, and original passages remain as background rather than undergoing
global name replacement.
\label{sec:gold-factorization}
For some relations, we also use several pieces of information
(\emph{evidence units}) intended to be combined
(Appendix~\ref{app:construction-plans}).

\paragraph{Add complete competing paths.}
\label{sec:competitor-chains}

For each reasoning step, we add an alternative with the same entity types
but a similar relation that does not meet the question's requirement.
We then add the subsequent facts needed to reach a different answer.
In the figure, Evan's son wrote Winter Letters, displayed at Harbor Museum;
his nephew wrote Cedar Pages, displayed at West Museum. Because both
manuscripts have museum evidence, a museum mention alone no longer selects
one path. The aim is to make the writer's relationship to Evan essential
for choosing between the answers.

We use two variants for constructing competitor paths, \emph{Flexible} and
\emph{Strict}. Flexible allows the
relation mismatch at the target step or, in a separate candidate path,
at a later step if this introduces no ambiguity or additional valid answer.
Strict allows only a target-step mismatch and keeps the entities in
unaffected dependencies unchanged. After problematic paths are removed,
at least one complete Strict path remains for each step
(Appendix~\ref{app:construction-competitors}); ablations compare both
variants in Section~\ref{sec:analysis}.

\subsection{Realize in Context}
\label{sec:distributed-placement}
\paragraph{Route facts into context.}
A \emph{host} is an existing passage that receives generated evidence.
Rewritten gold facts are inserted into the passages containing the original
supporting evidence; competing facts are assigned to model-selected compatible
passages. In the figure, $g_1,c_1$ go to Host A and $g_2,c_2$ to Host B.

\paragraph{Realization.}
\label{sec:context-realization}
Models turn the assigned facts and premises into sentences, such as
``Evan Hale's son, Liam Hale, wrote Winter Letters.'' Names and relations
must match the planned facts. The original text and its order are preserved,
without adding unplanned relations or explicitly ruling out competing answers.
We record the exact text expressing each fact and evidence unit.

\paragraph{Check the generated content.}
\label{sec:compiler-verification}
We first review the proposed facts and relations, then check whether the
text expresses them correctly. Automated checks identify missing facts,
changes to the original text, and incorrect evidence locations. Semantic
review checks names, relations, question meaning, unclear references, and
unintended correct answers. Problematic competing paths are removed, and
the remaining evidence and paths are checked against the construction
requirements.

Once the dataset is fixed, we test whether models answer correctly with
the full context (Full) and after removing specified evidence (Drop).
Deleting similar-length background text (Control) helps
distinguish evidence loss from general text removal
(Section~\ref{sec:audit-metrics}). Two annotators also assess sampled items
for quality (Appendix~\ref{app:human-spotchecks}).

\subsection{Adapt Across Settings}
\label{sec:construction-adapters}
\paragraph{\realhopmusique{}.}
MuSiQue provides annotated dependency graphs and supporting passages, which
define $G$ and the initial evidence hosts. We can therefore apply the
construction process directly to these annotations.

\paragraph{\realhopframes{}.}
FRAMES provides source links but no hop annotations.
We reconstruct the linked articles and ask a model to infer a linear
reasoning chain with quoted evidence for each step. We then apply the same
construction process.

\paragraph{\realhoplongbench{}.}
We start from a fictional graph; the source question guides only task
style. The same method constructs gold and competing paths, with evidence
placed at separate sentence boundaries across long-document chunks to test
retrieval and reasoning over longer contexts.
Across the three settings, the adapters vary how reasoning graphs are obtained
and evidence is placed while preserving the same objective: making each
reasoning step behaviorally necessary.

\section{Experimental Setup}
\label{sec:setup}

Our evaluation asks whether \realhop{} increases behavioral necessity, which
components create the challenge, and whether the long-context construction
separates models consistently across trials.

\paragraph{\realhopmusique{} and \realhopframes{} evaluations.}
\label{sec:setup-musique}
\label{sec:setup-frames}
\realhopmusique{} contains 790 source--constructed pairs spanning 2--4 hops;
the \realhopframes{} audit uses a selected cohort of 81 matched
source/constructed question pairs with inferred hop spans. Both evaluate
DeepSeek V4 Flash, Gemini 3.6 Flash, and Qwen 3.7 Max under Full and targeted
Drop and report Full accuracy and BNR.
Within each setting, the source and constructed sides use the same solver and
scoring protocol.
MuSiQue confidence intervals resample paired questions, retaining their
interventions and model responses together. FRAMES reports Avg@3 over its
recorded trials.

\paragraph{\realhoplongbench{} evaluation.}
\label{sec:setup-longbench}
\realhoplongbench{}, derived from LongBench v2~\citep{bai2024longbench2},
contains 216 open-ended short-answer questions with constructed evidence
chains. Each answer is a concise terminal entity or identifier; prompts range
from about 11k to 575k tokens.
We evaluate 16 models over three trials at each model's highest supported
reasoning effort. Exact normalized matches are accepted directly; remaining
answers use a frozen short-answer judge (Appendix~\ref{app:prompts}).
BNR uses the open-ended Full/Drop runs for three models reported in
Appendix~\ref{app:longbench-interventions}, not the three-trial Avg@3 scores.
As a format-matched reference, we evaluate both the source questions and
\realhoplongbench{} once under the official multiple-choice template with the
same 16 models; the constructed items use a frozen A--D order whose non-gold
choices are planned competitor endpoints
(Table~\ref{tab:longbench-three-trials}).

\section{Results}
\label{sec:results}

Section~\ref{sec:problem-and-audit} identified a substantial gap between
annotated multi-hop structure and behavioral necessity; the results below
show that \realhop{} sharply narrows this gap. We first test evidence
necessity through paired source--constructed comparisons in the MuSiQue and
FRAMES settings, then scale the same construction objective to long-context
multi-hop evaluation, where \realhoplongbench{} is assessed through targeted
intervention and repeated model comparisons.

\subsection{Behavioral Effects on \realhopmusique{}}
\label{sec:paired-necessity-results}

\begin{figure}[!t]
  \centering
  \IfFileExists{figures/figure2_full_edc.pdf}
    {\includegraphics[width=\linewidth]{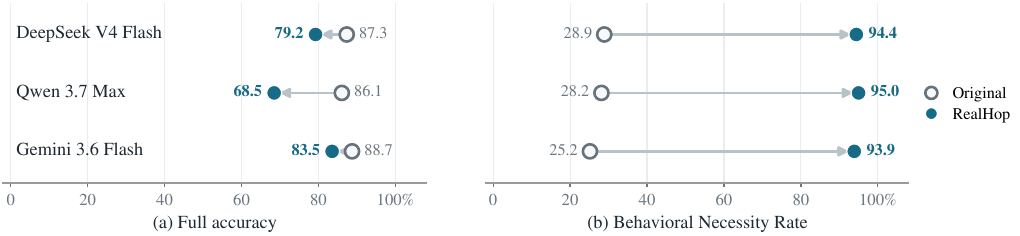}}
    {\includegraphics[width=\linewidth]{figure2_full_edc.pdf}}
  \caption{Full accuracy and BNR on source MuSiQue and \realhopmusique{}.
  Lower Full accuracy indicates greater difficulty;
  higher BNR indicates stronger behavioral dependence on the intended evidence.}
  \label{fig:full-bnr-main}
\end{figure}

Figure~\ref{fig:full-bnr-main} summarizes the paired source
MuSiQue/\realhopmusique{} comparison on all 790 questions.  Original Full
accuracy differs by only 2.6
points across the three models, offering little discrimination.  On
\realhopmusique{}, however, the between-model spread widens to 15.0 points, so
capability differences that stay compressed on the source items become easier
to see.

Despite this increase in difficulty, Full accuracy remains 68.5--83.5\%, providing a
substantial Full-correct population for BNR evaluation.  Within this harder
but still answerable setting, BNR rises from 25.2--28.9\% on Original to
93.9--95.0\% on \realhopmusique{}. The paired increases are 65.5 points for
DeepSeek (95\% CI 63.2--67.9), 68.7 for Gemini (66.4--71.0), and 66.8 for
Qwen (64.4--69.2); on the common-Full subset, gains remain 66.6--69.7 points
(Appendix~\ref{app:musique-common-full}). The construction therefore does more than lower accuracy:
it sharply increases behavioral dependence on the intended evidence. Its
observed BNR also substantially exceeds the levels in our FalseCoTQA and Plausible
Distractors audits (Section~\ref{sec:existing-benchmark-audit}), although those
use different benchmark samples.
Among incorrect Full answers, 64.3--69.5\% exactly match a planned competitor
endpoint (Appendix~\ref{app:competitor-hits}).

For six earlier models, paired accuracy drops by 25.3--47.9 points on
model-specific subsets of 305--353 questions, including 44.8 points for GPT-4o
(Appendix~\ref{app:historical-model-panel}). These declines establish increased
difficulty, but do not by themselves establish greater reliance on source
shortcuts: the subsets differ from the current-model panel, and these runs
do not measure BNR.

\subsection{\realhopframes{} Hop-Level Audit}
\label{sec:frames-results}

On the selected 81 matched FRAMES--\realhopframes{} pairs, each trial is
binarized and scored separately before averaging; each side's BNR conditions
on its own Full-correct subset.
Avg@3 source Full is 97.5--98.8\%, with BNR only
6.6--8.1\%. After construction, BNR rises to 88.9--89.9\% while Full remains
68.4--86.0\%, so the increase is not obtained by counting unsolved Full
questions as evidence-dependent successes. Per-trial scores appear in
Appendix~\ref{app:frames-bnr-trials}.

On the common-Full subset, BNR gains remain 81.8--83.5 points;
length-matched background scores exceed Drop
(Appendix~\ref{app:frames-matched-controls}).

\subsection{\realhoplongbench{} Across Repeated Trials}
\label{sec:longbench-three-trials}

Table~\ref{tab:longbench-three-trials} compares the two leaderboards.
The source leaderboard is comparatively compressed: accuracy under its
official protocol spans 58.3--72.2\%, and older models sometimes outrank newer
members of the same family. Gemini~3.5 Flash leads Gemini~3.6 Flash,
and Grok~4.5 leads Grok~4.6.

\begin{table}[H]
  \centering
  \footnotesize
  \setlength{\tabcolsep}{2.25pt}
  \caption{Source ranks use accuracy on the corresponding source questions
  under the official LongBench v2 multiple-choice template.
  Trials~1--3 and Avg@3 are open-ended scores on \realhoplongbench{}; the
  rightmost MCQ column evaluates the same constructed items under that
  template, run once. Eff. denotes the provider-specific reasoning-effort
  setting. Arrows give the rank change from the source leaderboard to Avg@3.
  Spread is the range of displayed scores across models.}
  \label{tab:longbench-three-trials}
  \renewcommand{\arraystretch}{1.05}
  \begin{tabular}{
    l
    c
    >{\columncolor{originalbg}}c
    >{\columncolor{originalbg}}c
    >{\columncolor{realhopbg}}c
    >{\columncolor{realhopbg}}c
    >{\columncolor{realhopbg}}c
    >{\columncolor{realhopsummarybg}}c
    >{\columncolor{realhopsummarybg}}c
    >{\columncolor{realhopsummarybg}}c
    >{\columncolor{realhopbg}}c}
    \toprule
    & & \multicolumn{2}{c}{
      \textcolor{rankneutral}{LongBench v2}}
    & \multicolumn{7}{c}{
      \textcolor{realhopblue}{\realhoplongbench{}}} \\
    \cmidrule(lr){3-4}\cmidrule(lr){5-11}
    Model & Eff. & \cellcolor{white}Acc. & \cellcolor{white}Rank
      & \cellcolor{white}Trial 1 & \cellcolor{white}Trial 2
      & \cellcolor{white}Trial 3 & \cellcolor{white}Avg@3
      & \cellcolor{white}Rank & \cellcolor{white}Change
      & \cellcolor{white}MCQ \\
    \midrule
    Claude Opus 5 & max & 70.4 & \ranksecond{} & 80.6 & 85.2 & 81.9 & 82.6 & \rankfirst{} & \textcolor{rankup}{$\uparrow 1$} & 94.4 \\
    GPT-5.6 Sol & xhigh & 68.1 & \rankthird{} & 63.9 & 63.9 & 65.7 & 64.5 & \ranksecond{} & \textcolor{rankup}{$\uparrow 1$} & 75.5 \\
    Kimi K3 & max & 65.7 & 6 & 52.8 & 59.7 & 50.9 & 54.5 & \rankthird{} & \textcolor{rankup}{$\uparrow 3$} & 64.8 \\
    Claude Opus 4.8 & max & 64.8 & 8 & 53.7 & 49.1 & 55.1 & 52.6 & 4 & \textcolor{rankup}{$\uparrow 4$} & 79.6 \\
    Grok 4.6 & xhigh & 65.3 & 7 & 43.1 & 40.3 & 44.0 & 42.4 & 5 & \textcolor{rankup}{$\uparrow 2$} & 60.2 \\
    Hy4 Preview & high & 64.4 & 10 & 39.8 & 43.1 & 41.7 & 41.5 & 6 & \textcolor{rankup}{$\uparrow 4$} & 52.8 \\
    Gemini 3.6 Flash & high & 68.1 & \rankthird{} & 41.2 & 39.8 & 42.6 & 41.2 & 7 & \textcolor{rankdown}{$\downarrow 4$} & 50.0 \\
    DeepSeek V4 Pro & max & 63.9 & 11 & 34.3 & 38.0 & 40.3 & 37.5 & 8 & \textcolor{rankup}{$\uparrow 3$} & 56.9 \\
    Qwen 3.8 Max & xhigh & 63.9 & 11 & 40.3 & 34.7 & 34.3 & 36.4 & 9 & \textcolor{rankup}{$\uparrow 2$} & 60.2 \\
    GLM 5.3 & max & 64.8 & 8 & 31.0 & 30.1 & 31.5 & 30.9 & 10 & \textcolor{rankdown}{$\downarrow 2$} & 53.2 \\
    DeepSeek V4 Flash & max & 63.0 & 13 & 32.4 & 31.0 & 27.3 & 30.2 & 11 & \textcolor{rankup}{$\uparrow 2$} & 55.6 \\
    Gemini 3.5 Flash & high & 72.2 & \rankfirst{} & 29.2 & 29.6 & 29.6 & 29.5 & 12 & \textcolor{rankdown}{$\downarrow 11$} & 38.9 \\
    GLM 5.2 & max & 61.6 & 15 & 25.0 & 27.8 & 28.7 & 27.2 & 13 & \textcolor{rankup}{$\uparrow 2$} & 50.9 \\
    Grok 4.5 & xhigh & 66.7 & 5 & 29.6 & 24.1 & 27.8 & 27.2 & 13 & \textcolor{rankdown}{$\downarrow 8$} & 48.6 \\
    Qwen 3.7 Max & xhigh & 62.5 & 14 & 26.9 & 24.1 & 28.2 & 26.4 & 15 & \textcolor{rankdown}{$\downarrow 1$} & 35.2 \\
    Kimi K2.6 & thinking & 58.3 & 16 & 27.3 & 27.8 & 23.1 & 26.1 & 16 & \textcolor{rankneutral}{--} & 49.5 \\
    \midrule
    Panel spread & & 13.9 & & 55.6 & 61.1 & 58.8 & 56.5 & & & 59.2 \\
    \bottomrule
  \end{tabular}
\end{table}

Avg@3 on \realhoplongbench{} instead spans 26.1--82.6\%. Across the three
runs, 15 of 16 models vary by no more than about six percentage points.
The constructed task therefore separates models more widely on long-context
multi-hop questions, including within-family comparisons that remain close on
the source leaderboard.
A same-format check rules out answer format as the sole cause of the wider
spread: under the identical MCQ template, \realhoplongbench{} spans 59.2
points versus 13.9 on the source questions. Its MCQ and open-ended Avg@3
rankings correlate strongly (Spearman $\rho=0.86$, $p<10^{-4}$).

An intervention check confirms that the intended evidence is behaviorally
consequential: BNR reaches 97.6\% for DeepSeek V4 Flash, 98.1\% for Gemini
3.6 Flash, and 96.5\% for Qwen 3.7 Max.

\section{Analysis}
\label{sec:analysis}

Section~\ref{sec:results} establishes the intended behavioral change. We now
isolate which parts of the construction method create this challenge.

We isolate the two competitor families on a stratified 80-question sample
spanning 2--4 hops. Manual review excludes seven items with ambiguous,
incomplete, or invalid questions or answer keys
(Appendix~\ref{app:ablation-details}). The remaining 73 items form the cohort
for the results below, reducing confounding from identified source-scoring
artifacts. All constructed conditions share the question and gold evidence:
\emph{Gold-only} contains no competitors, \emph{+Flexible} adds
less-constrained relation-level divergences, \emph{+Strict} adds
single-break paths with dependency inheritance, and \emph{Full} includes both.
The two middle conditions are parallel rather than cumulative: each adds one
competitor family to the Gold-only base, while Full combines them. This
separates family-specific effects from their joint effect.
\emph{Original} denotes the unmodified source questions used as a baseline. Six
earlier models characterize accuracy changes. Because their low Full accuracy
leaves few Full-correct questions, we estimate BNR with the three current
models. Each condition's BNR is conditioned on its own Full-correct subset and
is interpreted alongside that condition's Full accuracy.

\begin{table}[H]
  \centering
  \small
  \caption{Component ablation on a 73-item MuSiQue cohort (percent).
  \emph{Original} denotes source questions. Accuracy uses model-specific
  complete-case subsets; BNR averages hops within each question and is reported
  only for current models. Mean weights models equally.}
  \label{tab:component-ablation}
  \setlength{\tabcolsep}{4.5pt}
  \setlength{\aboverulesep}{0pt}
  \setlength{\belowrulesep}{0pt}
  \setlength{\extrarowheight}{2.2pt}
  \begin{tabular}{@{}lll
    >{\columncolor{originalbg}}c
    >{\columncolor{realhopbg}}c
    >{\columncolor{realhopbg}}c
    >{\columncolor{realhopbg}}c
    >{\columncolor{realhopbg}}c@{}}
    \toprule
    & & & \multicolumn{1}{c}{
      \textcolor{rankneutral}{MuSiQue}}
      & \multicolumn{4}{c}{
      \textcolor{realhopblue}{\realhopmusique{}}} \\
    \cmidrule(lr){4-4}\cmidrule(lr){5-8}
    Metric & Panel & Model
      & \cellcolor{white}Original
      & \cellcolor{white}Gold-only
      & \cellcolor{white}+Flexible
      & \cellcolor{white}+Strict
      & \cellcolor{white}Full \\
    \midrule
    \multirow{11}{*}{Accuracy}
      & \multirow{7}{*}{Earlier}
        & DeepSeek R1       & 84.9 & 72.6 & 57.5 & 42.5 & 37.0 \\
      & & Doubao 1.5 Pro    & 74.0 & 67.1 & 46.6 & 30.1 & 24.7 \\
      & & GPT-4o            & 71.2 & 68.5 & 31.5 & 24.7 & 16.4 \\
      & & Qwen Max          & 64.4 & 58.9 & 28.8 & 24.7 & 21.9 \\
      & & GLM-4-Plus        & 58.6 & 50.0 & 21.4 & 20.0 & 11.4 \\
      & & Qwen Turbo        & 39.7 & 35.6 & 19.2 &  9.6 &  5.5 \\
      \cmidrule(lr){3-8}
      & & Mean              & 65.5 & 58.8 & 34.2 & 25.3 & 19.5 \\
      \cmidrule(lr){2-8}
      & \multirow{4}{*}{Current}
        & DeepSeek V4 Flash & 97.3 & 94.5 & 93.2 & 89.0 & 83.6 \\
      & & Gemini 3.6 Flash  & 93.2 & 94.5 & 90.4 & 91.8 & 89.0 \\
      & & Qwen 3.7 Max      & 94.3 & 90.0 & 85.7 & 82.9 & 75.7 \\
      \cmidrule(lr){3-8}
      & & Mean              & 94.9 & 93.0 & 89.8 & 87.9 & 82.8 \\
    \midrule
    \multirow{4}{*}{BNR}
      & \multirow{4}{*}{Current}
        & DeepSeek V4 Flash & 22.7 & 56.2 & 86.5 & 86.6 & 91.7 \\
      & & Gemini 3.6 Flash  & 18.0 & 59.1 & 83.2 & 86.2 & 91.0 \\
      & & Qwen 3.7 Max      & 29.2 & 58.3 & 85.9 & 87.7 & 91.4 \\
      \cmidrule(lr){3-8}
      & & Mean              & 23.3 & 57.9 & 85.2 & 86.8 & 91.3 \\
    \bottomrule
  \end{tabular}
\end{table}

\paragraph{Difficulty.}
Table~\ref{tab:component-ablation} locates the difficulty in structured
competition rather than in rewriting the gold path.  Mean accuracy is only
modestly lower on Gold-only than on Original, then falls as competitor
families are added; competitor insertion accounts for 85.5\% of the
earlier-panel mean Original$\rightarrow$Full reduction. Strict paths cost more
than Flexible competitors overall, and both effects are far larger on the
earlier panel than on the current one.

\paragraph{Behavioral necessity.}
Difficulty alone does not establish the mechanism, so the BNR panel repeats the
ablation with targeted all-hop interventions. Rewriting the gold path more
than doubles panel-mean BNR. Adding either competitor family then raises BNR
by roughly 30 further points; combining both families yields an additional,
smaller gain.
These comparisons show stronger deletion sensitivity among each
condition's Full-correct items, which lower accuracy alone would not establish.

\section{Limitations}
\label{sec:limitations}

BNR measures whether a correct answer survives targeted deletion, not
whether the model internally used that hop. The estimate inherits the chosen
evidence unit, judge, and trial budget; Drop only indirectly reflects
processing, and most reported BNR uses one Full/Drop pair per condition. The
constructed items use synthetic entities
and planned chains inside source carriers, so they do not cover the full
range of natural information needs. On \realhoplongbench{}, the matched MCQ
check uses one trial and is not score-equivalent to open-ended Avg@3.
FRAMES controls use approximate matching with unverified trial counts;
common-Full results are subset-conditional.

\section{Conclusion}
\label{sec:conclusion}

We show that annotated multi-hop structure can remain behaviorally
unnecessary even when models answer correctly. \realhop{} addresses this gap
through a diagnose--construct--verify framework that audits claimed supports,
constructs complete competing paths, and validates evidence placement before
behavioral evaluation. We quantify behavioral dependence on intended evidence
with BNR, which measures how often targeted evidence removal prevents answer
recovery among Full-correct items. Panel-mean BNR rises from 27.4\% to 94.4\%
on \realhopmusique{}, reaches about 89\% on \realhopframes{} and above 96\% on
\realhoplongbench{}, while the long-context evaluation also yields a wider
model spread that persists across repeated trials. These results establish
evidence necessity as a benchmark design criterion rather than an assumption
inferred from annotations or answer accuracy.

\bibliographystyle{plainnat}
\bibliography{references}

\clearpage
\appendix
\startcontents[appendices]
\section*{Appendix Contents}
\printcontents[appendices]{}{1}{\setcounter{tocdepth}{2}}
\clearpage
\section{External Multi-Hop Audit Details}
\label{app:external-multihop-per-model}

The audit covers the 790 source MuSiQue questions and stratified 300-question
samples from HotpotQA and 2WikiMultiHopQA. MuSiQue Drop removes one gold
paragraph; HotpotQA and 2Wiki Drop remove the supporting sentences associated
with one title. We also audit stratified 300-item FalseCoTQA--MuSiQue and
300-item Plausible Distractors splits, the latter with two related
named-entity false-chain paragraphs. In both, Drop removes one official gold
support, false chains remain in all conditions, and Control deletes a
token-matched ordinary distractor. Table~\ref{tab:external-multihop-per-model}
gives the per-model results behind Table~\ref{tab:external-multihop-summary}.

\begin{table}[!t]
  \centering
  \small
  \caption{Per-model Full, Drop-one, Control, and BNR on the external
  multi-hop audit (percent).}
  \label{tab:external-multihop-per-model}
  \setlength{\tabcolsep}{4pt}
  \renewcommand{\arraystretch}{1.12}
  \begin{tabular}{@{}llcccc@{}}
    \toprule
    Benchmark & Model & Full & Drop-one & Control & BNR \\
    \midrule
    \multirow{3}{*}{MuSiQue}
      & DeepSeek V4 Flash & 87.3 & 64.1 & 86.9 & 28.9 \\
      & Gemini 3.6 Flash & 88.7 & 68.2 & 87.8 & 25.2 \\
      & Qwen 3.7 Max & 86.1 & 64.0 & 87.3 & 28.2 \\
    \midrule
    \multirow{3}{*}{HotpotQA}
      & DeepSeek V4 Flash & 80.7 & 68.8 & 81.3 & 17.2 \\
      & Gemini 3.6 Flash & 80.7 & 69.0 & 81.5 & 16.9 \\
      & Qwen 3.7 Max & 82.3 & 71.2 & 81.8 & 15.6 \\
    \midrule
    \multirow{3}{*}{2WikiMultiHopQA}
      & DeepSeek V4 Flash & 87.3 & 71.3 & 88.1 & 19.3 \\
      & Gemini 3.6 Flash & 89.7 & 75.6 & 90.0 & 17.0 \\
      & Qwen 3.7 Max & 88.7 & 76.6 & 89.3 & 14.8 \\
    \midrule
    \multirow{3}{*}{FalseCoTQA}
      & DeepSeek V4 Flash & 63.0 & 33.8 & 55.8 & 46.4 \\
      & Gemini 3.6 Flash & 64.3 & 36.3 & 62.1 & 44.8 \\
      & Qwen 3.7 Max & 64.0 & 38.9 & 61.9 & 39.3 \\
    \midrule
    \multirow{3}{*}{Plausible Distractors}
      & DeepSeek V4 Flash & 86.7 & 46.2 & 85.2 & 46.7 \\
      & Gemini 3.6 Flash & 88.3 & 44.8 & 85.8 & 49.2 \\
      & Qwen 3.7 Max & 88.3 & 43.5 & 88.2 & 50.8 \\
    \bottomrule
  \end{tabular}
\end{table}

\section{Matched-Background Specificity}
\label{app:matched-controls}

We retain matched-background deletion as a separate specificity diagnostic.
For each target support, the Control removes background spans that occur
verbatim in both Original and \realhopmusique{}, lie outside every gold support, and
exclude newly inserted competitor text.  The removed token count matches the
associated targeted Drop within five tokens or 5\%, whichever is larger.
Because a Control may span several locations, it matches deletion volume
rather than discourse position.

\begin{table}[!t]
  \centering
  \small
  \caption{Matched-background specificity on aligned MuSiQue support
  deletions (percent), computed on Full-correct items with valid Drop and
  Control results.}
  \label{tab:appendix-matched-controls}
  \begin{tabular}{lcccc}
    \toprule
    & \multicolumn{2}{c}{Drop survival}
      & \multicolumn{2}{c}{Control survival} \\
    \cmidrule(lr){2-3}\cmidrule(lr){4-5}
    Model & Original & \realhopmusique{} & Original & \realhopmusique{} \\
    \midrule
    DeepSeek V4 Flash & 73.4 & 5.7 & 96.6 & 93.0 \\
    Qwen 3.7 Max      & 74.4 & 5.8 & 97.8 & 89.7 \\
    Gemini 3.6 Flash  & 76.3 & 6.2 & 97.2 & 92.8 \\
    \bottomrule
  \end{tabular}
\end{table}

Control survival remains 89.7--93.0\% on \realhopmusique{}, while targeted Drop
survival is 5.7--6.2\%.  Thus the BNR increase is not accompanied by a comparable
loss under generic background deletion.

\section{MuSiQue Common-Full Analysis}
\label{app:musique-common-full}

Each side's BNR in Section~\ref{sec:paired-necessity-results} conditions on
its own Full-correct questions, so the Original and \realhopmusique{}
populations differ. This check recomputes both sides on the same questions
from the saved responses; no additional model calls are made.

Let $\mathcal I^{\cap}=\{i:F_i^{O}=F_i^{R}=1\}$, restricted to questions with
complete Drop evaluations on both sides. We recompute both sides' BNR on this
set with Equation~\ref{eq:bnr}, averaging hops within each question and then
questions equally.

\begin{table}[!t]
  \centering
  \small
  \setlength{\tabcolsep}{3.5pt}
  \caption{MuSiQue BNR on questions with Full correctness on both sides.
  BNR is percent; the paired increase and its interval are percentage points.}
  \label{tab:musique-common-full}
  \begin{tabular}{@{}lcccc@{}}
    \toprule
    Model & $|\mathcal I^{\cap}|$ & BNR Original & BNR \realhopmusique{}
      & Increase [95\% CI] \\
    \midrule
    DeepSeek V4 Flash & 559 & 28.04 & 94.66 & 66.6 [64.0, 69.1] \\
    Gemini 3.6 Flash  & 599 & 24.36 & 94.05 & 69.7 [67.2, 72.2] \\
    Qwen 3.7 Max      & 482 & 27.35 & 94.90 & 67.5 [64.8, 70.3] \\
    \bottomrule
  \end{tabular}
\end{table}

The intersection retains 559, 599, and 482 of the 790 questions. BNR
increases remain 66.6--69.7 points (Table~\ref{tab:musique-common-full}),
compared with 65.5--68.7 points on each side's own Full-correct population.
Restricting to the common set changes the increase by +1.10, +0.98, and
+0.74 points, respectively, with paired 95\% intervals of $[0.02,2.22]$,
$[0.00,1.97]$, and $[-0.64,2.18]$. The result supports persistence on the
jointly answerable subset, not equivalence of populations or replacement of
the full-cohort results. Intervals use 10{,}000 paired whole-question
bootstrap resamples with percentile endpoints, base seed 20260925, and fixed
model offsets; each resampled question
retains its Full and Drop results on both sides, and both Full-correct
definitions are recomputed within every resample.

\section{Competitor-Endpoint Error Analysis}
\label{app:competitor-hits}

Table~\ref{tab:competitor-hits} reports how often an incorrect Full response
on the 790 \realhopmusique{} questions exactly matches the terminal answer of
a planned competitor.  Full correctness uses the same responses and
denominator as Figure~\ref{fig:full-bnr-main}.

\begin{table}[!t]
  \caption{Planned competitor-endpoint matches among Full answers on the 790
  \realhopmusique{} questions (percent).}
  \label{tab:competitor-hits}
  \centering
  \small
  \begin{tabular}{lcc}
    \toprule
    Model & Wrong hit & Share of errors \\
    \midrule
    DeepSeek V4 Flash & 14.4 & 69.5 \\
    Gemini 3.6 Flash  & 11.1 & 67.7 \\
    Qwen 3.7 Max      & 20.3 & 64.3 \\
    \bottomrule
  \end{tabular}
\end{table}

Exact competitor endpoints account for 64.3--69.5\% of incorrect Full
answers across models (wrong-hit rates: 11.1--20.3\%). Exact matching excludes
paraphrases and other branch-induced errors, making this a conservative
signature that models select the alternatives introduced by construction.

\section{Historical Model Difficulty}
\label{app:historical-model-panel}

Table~\ref{tab:historical-model-panel} reports the complete paired results
summarized in Section~\ref{sec:paired-necessity-results}. This earlier model
panel measures the difficulty shift under \realhopmusique{}, not behavioral
necessity.

\begin{table}[!t]
  \centering
  \small
  \caption{Paired accuracy on source MuSiQue and the corresponding
  \realhopmusique{} questions for six earlier models (percent).}
  \label{tab:historical-model-panel}
  \begin{tabular}{lcccc}
    \toprule
    Model & Paired $n$ & Original & \realhopmusique{} & $\Delta$ \\
    \midrule
    GPT-4o          & 353 & 62.6 & 17.8 & $-44.8$ \\
    Qwen Turbo      & 352 & 33.8 &  8.5 & $-25.3$ \\
    Qwen Max        & 352 & 55.1 & 15.6 & $-39.5$ \\
    GLM-4-Plus      & 337 & 51.9 & 15.1 & $-36.8$ \\
    Doubao 1.5 Pro  & 340 & 64.7 & 20.9 & $-43.8$ \\
    DeepSeek R1     & 305 & 79.3 & 31.5 & $-47.9$ \\
    \bottomrule
  \end{tabular}
\end{table}

All six models decline: mean paired accuracy falls from 57.9\% on source
MuSiQue to 18.2\% on \realhopmusique{}, a 39.7-point reduction. DeepSeek R1
remains strongest in both conditions but has the largest decrease. The effect
is therefore neither model-specific nor a rank reversal. These runs measure
difficulty only, not BNR.

\section{Component Ablation Details}
\label{app:ablation-details}

After the removals described below, the stratified sample contains 39 two-hop,
23 three-hop, and 11 four-hop questions.  Flexible competitors significantly
reduce paired accuracy for GPT-4o ($-37.0$ points, $p<10^{-5}$), Qwen Max
($-30.1$, $p<10^{-5}$), GLM-4-Plus ($-28.6$, $p<10^{-4}$), Doubao ($-20.5$,
$p=0.008$), and Qwen Turbo ($-16.4$, $p=0.008$); the DeepSeek R1 reduction is
borderline ($-15.1$, $p=0.061$).  Positive interaction terms of 9.6--28.7
points indicate overlapping rather than additive failure mechanisms.

\paragraph{Removed source questions.}
Seven source items with defective questions or keys are excluded
(Table~\ref{tab:ablation-removed-items}). Candidates were items that at most
two of nine models answered correctly on source MuSiQue, then checked by hand
against the reference key.

\begin{table}[!t]
  \centering
  \footnotesize
  \caption{The seven source questions removed from the component ablation.}
  \label{tab:ablation-removed-items}
  \renewcommand{\arraystretch}{1.25}
  \begin{tabular}{@{}p{0.30\linewidth}p{0.20\linewidth}p{0.44\linewidth}@{}}
    \toprule
    Question & Reference key & Defect \\
    \midrule
    Who is the child of Mahmoud Mirza's father?
      & Ahmad Shah Qajar
      & Self-referential: Mahmoud Mirza is himself a child of that father, and
        all nine models answer with his name. \\
    What team does the winner of the 2017 BBC African Footballer of the Year
    play for?
      & Egypt national football team
      & Club and national team are both valid; models answer with the club. \\
    What mountain can you see from Portland, in the state that Raven Creek is
    located in?
      & Tualatin Mountains
      & Several mountains are visible from Portland; the key records one. \\
    Bancroft's county borders what county?
      & Haliburton County
      & Multiple bordering counties; answers that name the key alongside
        another county are scored incorrect. \\
    When did Nissan, the Acura Legend maker and the Scion owner open US
    assembly plants?
      & 1981
      & Models answer ``early 1980s'', which the key cannot accept. \\
    When did the luxury division of the employer of Katsuaki Watanabe change
    the body style of the rx 350?
      & Sales began worldwide in April 2012
      & Key is a sentence rather than a date; models answer ``March 2012''. \\
    How were people from whom new coins were a proclamation of independence by
    the Somali Muslim Ajuran Empire expelled from the natural boundary between
    Thailand and Vat Yotkeo's country?
      & The dynasty regrouped and defeated the Portuguese
      & Four-hop composition is not interpretable as a single question. \\
    \bottomrule
  \end{tabular}
\end{table}

BNR follows Equation~\ref{eq:bnr}: each Full-correct question with complete
Drop evaluations contributes the mean over its gold supports, and questions
are averaged. The 73-item sample has 191 gold supports
($39\times 2+23\times 3+11\times 4$).

\FloatBarrier
\section{\realhoplongbench{} Targeted Interventions}
\label{app:longbench-interventions}

Full is Trial~1 in Table~\ref{tab:longbench-three-trials}. BNR follows
Equation~\ref{eq:bnr}: each Full-correct question with complete targeted Drop
evaluations contributes the mean over its targeted hops, and questions are
averaged.

\begin{table}[!t]
  \centering
  \small
  \caption{Targeted-hop intervention results on \realhoplongbench{} (percent).}
  \label{tab:longbench-targeted-interventions}
  \begin{tabular}{lcc}
    \toprule
    Model & Full & BNR [95\% CI] \\
    \midrule
    DeepSeek V4 Flash & 32.4 & 97.6 [94.3, 100.0] \\
    Gemini 3.6 Flash & 41.2 & 98.1 [95.5, 100.0] \\
    Qwen 3.7 Max & 26.9 & 96.5 [93.1, 99.4] \\
    \bottomrule
  \end{tabular}
\end{table}

\section{Random-Window Diagnostics}
\label{app:random-window-probe}

Three frozen contiguous windows each remove 20\% of the context. The probe
measures spatial sensitivity, not hop necessity. Constructed chains are
placed across long carriers, so random deletion can change accuracy.

\paragraph{Protocol.}
For existing long-context sets, we sample 100 questions each from LongBench
v2~\citep{bai2024longbench2}, NarrativeQA~\citep{bai2024longbench,kocisky2017narrativeqa},
and InfiniteBench LongBookQA~\citep{zhang2024infinitebench}.  Each Full input
is paired with three shared variants that remove a uniformly positioned
contiguous 20\% window.  LongBench v2 uses option matching; the open-ended
tasks use a common short-answer judge.  Paired effects use complete cases.
On the matched 216-question \realhoplongbench{} panel, three frozen circular
masks each remove a contiguous 20\% token window; accuracy uses the fixed
216-question denominator, and Full is Trial~1 in
Table~\ref{tab:longbench-three-trials}.
We evaluate Loong~\citep{wang2024loong} using the same Full plus three-mask
schedule; its open-ended responses use Loong's official 1--100 judge prompt.

On LongBench~v2, NarrativeQA, and LongBookQA, panel-mean paired accuracy
falls by only 1.9--2.7 points
(Table~\ref{tab:appendix-random-mask-full}).  Across the nine
model--benchmark pairs, 55.3--83.3\% of Full-correct items remain correct
under all three masks.  Several paired confidence intervals include zero,
so the net drop is not a necessity estimate.

\begin{figure}[!t]
  \centering
  \includegraphics[width=0.92\linewidth]{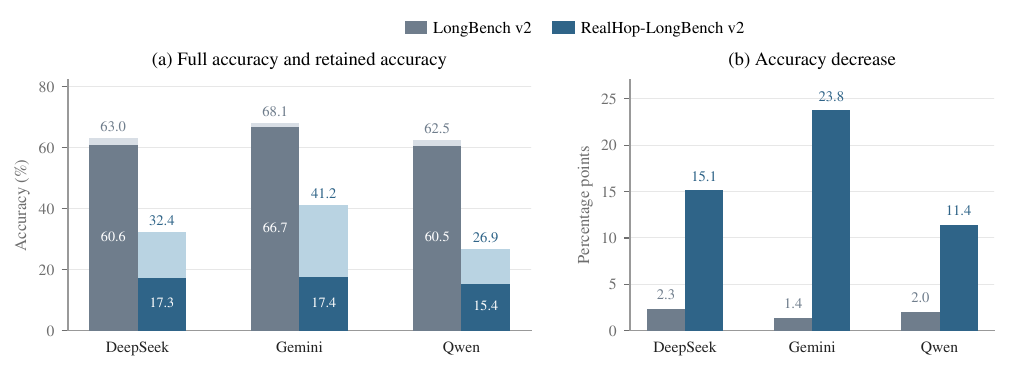}
  \caption{Sensitivity to three frozen circular 20\% masks on the matched
  216-question panel.
  \textbf{(a)} Full accuracy decomposed into accuracy retained after Mask20
  (dark lower segment) and the decrease (light upper segment).
  \textbf{(b)} Mean accuracy decrease.
  Each Mask20 value averages three frozen masks per question.}
  \label{fig:mask20-sensitivity}
\end{figure}

On the matched 216 IDs, the same protocol lowers source LongBench v2 accuracy
by 1.4--2.3 points, but lowers \realhoplongbench{} accuracy by 11.4--23.8
points (Figure~\ref{fig:mask20-sensitivity},
Table~\ref{tab:longbench-random-mask}).  This gap is consistent with
distributing constructed evidence across the carrier.  Which hops are
necessary is given by the targeted Drop results in
Appendix~\ref{app:longbench-interventions}.

\begin{table}[!t]
  \centering
  \scriptsize
  \caption{Random 20\% mask results on \realhoplongbench{} (percent).}
  \label{tab:longbench-random-mask}
  \begin{tabular}{lccccccc}
    \toprule
    Model & Full & Mask 1 & Mask 2 & Mask 3 & Mask avg.
      & Accuracy decrease [95\% CI] & Full-correct survival \\
    \midrule
    DeepSeek V4 Flash & 32.4 & 18.5 & 16.7 & 16.7 & 17.3
      & 15.1 [8.8, 21.6] & 24.3 \\
    Gemini 3.6 Flash & 41.2 & 19.4 & 17.1 & 15.7 & 17.4
      & 23.8 [16.8, 30.6] & 21.0 \\
    Qwen 3.7 Max & 26.9 & 17.6 & 11.6 & 17.1 & 15.4
      & 11.4 [5.2, 17.7] & 20.1 \\
    \bottomrule
  \end{tabular}
\end{table}

\begin{table}[!t]
  \centering
  \scriptsize
  \setlength{\tabcolsep}{3pt}
  \caption{Full and random-mask accuracy on four long-context audits
  (percent).}
  \label{tab:appendix-random-mask-full}
  \resizebox{\linewidth}{!}{%
  \begin{tabular}{llccccc}
    \toprule
    Benchmark & Model & Complete $n$ & Full fixed & Mask fixed
      & $\Delta$ fixed & $\Delta$ paired [95\% CI] \\
    \midrule
    LongBench v2 & DeepSeek & 100 & 73.0 & 68.7 & $-4.3$ & $-4.3$ [$-10.0,1.7$] \\
                 & Gemini  & 100 & 71.0 & 69.0 & $-2.0$ & $-2.0$ [$-6.3,2.7$] \\
                 & Qwen     & 94 & 69.0 & 65.3 & $-3.7$ & $-1.8$ [$-7.8,4.3$] \\
    \midrule
    NarrativeQA & DeepSeek & 100 & 57.0 & 57.0 & $\phantom{-}0.0$ & $\phantom{-}0.0$ [$-6.7,6.7$] \\
                & Gemini  & 100 & 54.0 & 54.3 & $+0.3$ & $+0.3$ [$-5.3,6.3$] \\
                & Qwen     & 90 & 47.0 & 44.7 & $-2.3$ & $-5.9$ [$-13.0,1.1$] \\
    \midrule
    LongBookQA & DeepSeek & 98 & 81.0 & 76.3 & $-4.7$ & $-4.4$ [$-10.2,1.4$] \\
               & Gemini  & 100 & 82.0 & 77.7 & $-4.3$ & $-4.3$ [$-8.7,0.0$] \\
               & Qwen     & 76 & 58.0 & 62.7 & $+4.7$ & $+0.9$ [$-4.4,6.1$] \\
    \midrule
    Loong & DeepSeek & 100 & 81.0 & 63.3 & $-17.7$ & $-17.7$ [$-23.3,-12.3$] \\
          & Gemini  & 100 & 80.0 & 63.0 & $-17.0$ & $-17.0$ [$-23.0,-11.0$] \\
          & Qwen     & 100 & 80.0 & 58.0 & $-22.0$ & $-22.0$ [$-28.7,-15.7$] \\
    \bottomrule
  \end{tabular}
  }
\end{table}

\FloatBarrier
\paragraph{Why the LongBench v2 net drop is small.}

If a question depended on one indispensable location and mask starts were
uniform, each trial would remove that location with probability 20\%, and
survival under all three masks would be $0.8^3=51.2\%$. Observed all-mask
survival among Full-correct questions is 75.3\% for DeepSeek, 81.7\% for
Gemini, and 75.4\% for Qwen.

Net accuracy also hides two-way flips. Conditional on a Full-correct
opportunity, mask trials become incorrect at 12.8\%, 8.0\%, and 12.3\%,
still below the 20\% single-point prediction. Reverse flips, where an
incorrect Full answer becomes correct, cancel 3.7--6.7 points of that
one-sided loss. Many supporting spans are therefore not behaviorally unique.

Loong falls by 17.0--22.0 points under the same protocol, so a small Mask20
effect is not a general property of long-context benchmarks.

\section{Evidence Construction Details}
\label{app:construction-details}

The details below describe the MuSiQue construction procedure and its
adaptation to FRAMES.

\subsection{Plans and Traceability}
\label{app:construction-plans}

The planner emits a structured record containing a shared entity registry,
question substitutions, hop dependencies, and one gold realization plan per
hop. Each realization plan contains evidence units with stable identifiers,
atomic facts, and a composition rule. The planner requires at least
$\lceil |H|/2\rceil$ hardening targets, including the final hop; each selected
target must have at least two planned evidence units.
For example, rather than directly stating that Winter Letters is displayed
at Harbor Museum, the plan uses two complementary premises: ``Winter Letters
is listed under entry E7 in the exhibition catalogue'' and ``Entry E7 names
Harbor Museum as the display venue.'' Together they link the manuscript to
the museum; neither premise alone establishes that relation. These units
support the same hop in the original question graph, so the annotated hop
count is unchanged. Each unit is linked to its corresponding text.

Later records keep the link from a graph fact to its text:
\begin{itemize}
    \item \textbf{Competitor plan:} branch and fact identifiers, the protected
    hop, relation status, entity bindings, and host exclusions.
    \item \textbf{Route plan:} retained fact identifiers and assigned host
    passages, linked to the compiled fact plan.
    \item \textbf{Realization:} one record per planned source identifier,
    with realized text spans and, for gold hops, individual evidence-unit
    spans.
\end{itemize}
The main MuSiQue evaluation removes a supporting paragraph, not each
premise within it.

\subsection{Flexible and Strict Competitor Contracts}
\label{app:construction-competitors}

The \emph{protected hop} $h$ indexes the support-deletion target, not a hop
exempt from deletion; the \emph{break hop} $b$ locates the relation mismatch.
The full pipeline targets every question hop, independently of the gold
premise-decomposition subset and without using behavioral Drop outcomes.

\textbf{Flexible.} The planner proposes a non-entailing relation at $b=h$
with a complete continuation. Where safe, it also proposes a separate path
with $b\in\operatorname{Desc}_G(h)$: the protected relation matches, but a
later dependency does not. Prefixes that introduce ambiguous referents or
another valid answer are rejected. Fact-level pruning may leave fragments.

\textbf{Strict.} The sole mismatch must satisfy $b=h$, using a type-compatible,
non-entailing predicate. Only $R_h=\{h\}\cup\operatorname{Desc}_G(h)$ is
regenerated; other dependencies inherit exact gold bindings. Descendants
preserve required relations but may use alternative entities. Unsafe paths
are removed atomically. A change of narrative framing alone is not a break;
neither mode labels competitors as wrong.

\textbf{Example.} Consider kinship, authorship, and display as separate hops
in Figure~\ref{fig:realhop-construction}, and target authorship. A Strict
candidate can inherit Liam's kinship, state that he \emph{translated} another
manuscript, and give its display venue: translation does not entail authorship,
while display remains matched. Flexible also allows this pattern. A separate
Flexible proposal could instead match authorship but replace \emph{displayed}
with \emph{stored}. Strict excludes this downstream break for the same target.
Flexible still rejects it if the added manuscript makes the question's
referent ambiguous. These hypothetical plans illustrate constraints, not
extra facts in the figure. No downstream-break option exists at a terminal hop.

\textbf{Coverage and ablations.} Two independent Strict candidates with
distinct entities and answers are proposed per hop; at least one complete
path per hop must survive pruning. Flexible fragments cannot fill missing
Strict coverage. Full combines both families; +Flexible and +Strict add each
separately to Gold-only. Removing Strict in an ablation removes its coverage
requirement. These are construction contracts, not guarantees of behavioral
necessity; routing and acceptance checks follow below.

\subsection{Routing, Realization, and Validation Boundaries}
\label{app:construction-validation}

A route must respect each fact's explicit forbidden-passage list and account
for every retained fact. A host is one dataset passage, not necessarily an
entire source document. Flexible plans must include the protected support in
each competing fact's forbidden list. In the v8.6.2 backend, Strict plans
are validated separately and do not universally enforce that inclusion, so
Strict facts may share the protected gold-support passage. A Strict path
with at least two generated facts uses at least two hosts, and its near-miss
host cannot also contain its generated exact continuation. Base chains with
at least three facts also use at least two hosts. These are host-level rules,
not a universal token-distance requirement. MuSiQue Drop removes the entire
rewritten support passage, including any co-located gold and competitor text.
Neither generated nor inherited competitor evidence is therefore guaranteed
to survive a deletion; planned-chain completeness is not a post-deletion
survival check.

The realizer retains the original title and text in order, adding complete
sentences at safe boundaries. It must not strengthen the planned claim's scope
or create an unplanned conclusion. More broadly, document organization can
matter beyond local rewriting alone \citep{tao2026beyond}. Mechanical checks cover source
identifiers, planned-unit coverage, and exact span occurrence. If
original-text preservation fails, a repair reconstructs the source body and
appends recovered additions; anaphora repair can likewise move additions.
Generated entities do not inherit unstated properties of the original
entities in the retained source text.

Blocking structural checks include graph linkage, entity-registry
consistency, executable evidence plans, route coverage, and realization
correspondence. The semantic audit rejects material question-contract
violations, ambiguous bindings, and unintended complete answer paths.
Salience and directional-hardening scores are recorded separately. In
v8.6.2, missing unit identifiers or invalid text-span mappings block acceptance,
but premise-separation and directional-hardening issues are non-fatal
diagnostics. Structural coverage therefore does not certify that every
realized premise is complementary or behaviorally necessary.
Human spot checks of gold--question match and competitor non-entailment
are reported in Appendix~\ref{app:musique-spotcheck} and
Appendix~\ref{app:longbench-spotcheck}.

After realization, competitor pruning is limited to two iterations. Each
removal is followed by a coverage check, and items that lose required
coverage are excluded. Full/Drop outcomes are not used for retention;
behavioral auditing begins only after construction and validation.

\subsection{FRAMES Adapter and Hop-Level Audit}
\label{app:frames-construction}

FRAMES supplies questions and Wikipedia links, not hop labels. We assemble
each article from those links. When tables are kept, HTML rows that overlap
the question or the reference answer are flattened and appended in source
order. Hop proposal, realization, and source-side evaluation use this same
text.

A model then proposes a linear chain: each hop quotes a span in a source
passage, and each later hop points to the previous one. Several hops may
quote the same article. Checks confirm that the fields are filled and that
each quotation occurs in the assembled text. After this conversion, the
same MuSiQue construction procedure generates the gold and competitor text. Only
entity-answer plans are realized: numbers are not rewritten and derived
answers are not recomputed. Two items with stale values are dropped,
leaving 81 questions and 208 hops per side. Article bodies are capped at
24{,}000 characters, plus up to 1{,}500 characters of selected table rows.
This is not the full FRAMES test set.

For the reported BNR analysis, DeepSeek V4 Flash, Gemini 3.6 Flash, and
Qwen 3.7 Max answer source and constructed items under Full and every
single-hop Drop. The reported analysis uses three trials per condition.
DeepSeek V4 Pro
scores each response as 1, 0.5, or 0 from the question, reference, and
answer, without the context (Appendix~\ref{app:prompts}).

Each Drop removes one inferred hop: all recorded evidence-unit spans of
that hop on the constructed side, or the quoted evidence span on the source
side. Deletion follows the exact span, then removes string-matched
endpoint cooccurrences. This cleanup may leave residual paths or remove
incidental cooccurrences. A hop's units are deleted
jointly, so this audit does not test each split premise independently.
Supplementary matched-background and common-Full analyses are reported in
Appendix~\ref{app:frames-matched-controls}; the background control is separate
from the BNR calculation.

\paragraph{BNR recomputation from recorded trials.}
\label{app:frames-bnr-trials}
We recompute Equation~\ref{eq:bnr} from the recorded per-condition trial
scores. A score of 1 is correct; scores of 0.5 and 0 are incorrect.
We require every expected
hop condition, matched to the intervention audit, and all three Full/Drop
scores to be present. The available complete panels contain 80, 81, and 79
source--constructed pairs for DeepSeek, Gemini, and Qwen, respectively,
covering 205, 208, and 202 hop conditions per side. The remaining 1, 0, and 2
items have incomplete records and are excluded. Only complete three-trial
panels are retained.

For each trial $t$ and side $s$, define
$F_i^{s,t}=\mathbf{1}[\mathrm{score}_{i,\mathrm{Full}}^{s,t}=1]$ and
$D_{ih}^{s,t}=\mathbf{1}[\mathrm{score}_{ih,\mathrm{Drop}}^{s,t}=1]$.
On the complete-panel population, we first select
$\mathcal I_{s,t}^{+}=\{i:F_i^{s,t}=1\}$, then average $1-D_{ih}^{s,t}$ over
hops within each selected item and over items. Full accuracy uses the
complete-pair denominator, whereas BNR uses $|\mathcal I_{s,t}^{+}|$.
The main summary is the arithmetic mean of the three trial-level Full
accuracy and BNR values reported below. No majority vote, best-trial
selection, or pre-binarization score average is used. Each side has its own
Full-correct subset, so the reported BNR differences are not restricted to
items answered correctly on both sides. The same metric is used for MuSiQue,
but the intervention, judge configuration, and complete-case population
remain distinct.

\begin{table}[!t]
  \centering
  \small
  \caption{FRAMES BNR recomputed independently for each recorded trial
  (percent). O/R denote source/constructed items; $n^+$ is the corresponding
  Full-correct question count. Complete-panel counts are 80, 81, and 79
  throughout. Avg@3 Full and BNR in Section~\ref{sec:frames-results} are
  the arithmetic mean of these trial-level metrics.}
  \label{tab:frames-bnr-trials}
  \begin{tabular}{@{}lrcccc@{}}
    \toprule
    \multirow{2}{*}{Model} & \multirow{2}{*}{Trial}
      & \multicolumn{2}{c}{$n^+$} & \multicolumn{2}{c}{BNR} \\
    \cmidrule(lr){3-4}\cmidrule(lr){5-6}
    & & O & R & O & R \\
    \midrule
    \multirow{3}{*}{DeepSeek V4 Flash}
      & 1 & 79 & 59 & 7.81 & 89.83 \\
      & 2 & 79 & 65 & 6.75 & 88.03 \\
      & 3 & 79 & 62 & 8.44 & 88.87 \\
    \midrule
    \multirow{3}{*}{Gemini 3.6 Flash}
      & 1 & 80 & 71 & 5.21 & 89.18 \\
      & 2 & 79 & 69 & 7.59 & 87.78 \\
      & 3 & 80 & 69 & 7.08 & 91.09 \\
    \midrule
    \multirow{3}{*}{Qwen 3.7 Max}
      & 1 & 77 & 51 & 6.93 & 89.87 \\
      & 2 & 76 & 58 & 8.33 & 88.62 \\
      & 3 & 78 & 53 & 9.19 & 91.19 \\
    \bottomrule
  \end{tabular}
\end{table}

\subsection{FRAMES Matched Controls and Sensitivity}
\label{app:frames-matched-controls}

Two supplementary checks address different confounds: changes in the
Full-correct population and nonspecific damage from deleting text. All
analyses below reuse saved scores; no additional model calls are made.
Records are deduplicated by retaining the latest entry before filtering for
completeness. The 80/81/79 complete source--constructed panels are unchanged.

\paragraph{Common-Full subset.}
For trial $t$, let $\mathcal I_t^{\cap}=\{i:F_i^{O,t}=F_i^{R,t}=1\}$.
We recompute both sides' BNR on this same set, retaining the main paper's
rule that only score 1 is correct; a Drop score of 0.5 is therefore a failure.
We average hops within each question, questions within each trial, and then
the three trial estimates equally. Trial indices align the recorded attempts,
not shared random draws. This differs from pooling all retained item--trials
or applying a threshold to the three-trial average score.

\begin{table}[!t]
  \centering
  \small
  \setlength{\tabcolsep}{3.5pt}
  \caption{FRAMES BNR on trials with Full correctness on both sides.
  Counts give $|\mathcal I_t^{\cap}|$ for trials 1/2/3; O/R are source/constructed.
  BNR is percent; the paired increase and its interval are percentage points.}
  \label{tab:frames-common-full}
  \begin{tabular}{@{}lcccc@{}}
    \toprule
    Model & Counts & BNR O & BNR R & Increase [95\% CI] \\
    \midrule
    DeepSeek V4 Flash & 59/65/61 & 6.80 & 88.85 & 82.1 [75.2, 88.2] \\
    Gemini 3.6 Flash  & 70/67/68 & 5.75 & 89.30 & 83.5 [77.4, 89.1] \\
    Qwen 3.7 Max      & 49/56/52 & 7.79 & 89.57 & 81.8 [74.7, 88.2] \\
    \bottomrule
  \end{tabular}
\end{table}

The intersection retains 72, 71, and 61 distinct questions, contributing
185, 205, and 157 item--trials. The joint Full criterion excludes 55, 38, and
80 item--trials, respectively. BNR increases remain 81.8--83.5 points
(Table~\ref{tab:frames-common-full}).

Relative to each side's own Full-correct population, restricting to the common
set changes the increase by +0.81, +0.83, and +0.03 points, respectively,
with paired 95\% intervals of $[-1.03,2.99]$, $[-0.62,2.62]$, and $[-2.39,2.54]$.
These intervals directly estimate the change in effect, rather than comparing
the overlap of separate intervals. The result supports persistence on the
jointly answerable subset, not equivalence of populations or replacement of
the full-cohort results.

\paragraph{Background control and condition pairing.}
The separately run background control selects
background text using lexical protection of all recorded hop evidence,
subjects, answers, and, on the constructed side, competitor spans. Its budget
is the summed character length of the target hop's recorded spans, with a
25\% tolerance and an attempt to follow their passage distribution; it is
not a token-level or exact-position match. This budget can differ from the
actual Drop removal, which includes repeated occurrences and relation-closure
cleanup. Short-span cases may use a flagged sentence-fragment fallback.

We align Control and Drop by sample ID and the explicit hop index. Control
means are stored as a list: we recover its condition names using the producer's
lexicographic order of feasible conditions, requiring exact per-item agreement
between metadata, score-list length, and missing-condition counts. We do not
shift list positions across infeasible conditions. Every selected comparison
uses the same question--hop conditions for Full, Control, and Drop, averaging
condition means within a question and then questions equally. These are mean
judge scores on a 0/0.5/1 scale, not binary accuracy or BNR. Full and Drop have
three recorded scores per condition; control trial counts are not committed,
so their three-trial completeness cannot be verified or reconstructed from
an average score.

Constructed-side control coverage is 204/205, 207/208, and 201/202 conditions;
all feasible constructed-side deletions are sentence-aligned. Original-side
coverage is 198/205, 201/208, and 196/202, with nine feasible fragment deletions
per model; these are excluded from sentence-only sensitivity checks. Missing
controls are omitted together with their paired Drop conditions, not counted
as failures. On all feasible constructed-side pairs, Control--Full mean-score
differences are +0.59, -2.19, and -0.58 points, with paired 95\% intervals
[-3.68, 5.14], [-4.42, -0.13], and [-5.51, 4.23]. Thus we do not claim that
background deletion has exactly zero effect, even though its damage is much
smaller than the targeted-Drop effect.

\paragraph{Actual-length sensitivity.}
Using deletion metadata alone, we additionally require a whole-sentence
control and $|L^K_{ih}/L^D_{ih}-1|\leq\epsilon$, where $L^K_{ih}$ is the
control's removed character count and $L^D_{ih}$ is the actual Full--Drop
character difference, including closure. We report all three nested bands,
$\epsilon\in\{0.25,0.10,0.05\}$, without selecting conditions by model scores.
This is an offline, post-hoc robustness check, not a new set of model runs.

\begin{table}[!t]
  \centering
  \footnotesize
  \setlength{\tabcolsep}{3pt}
  \caption{Constructed FRAMES background-deletion sensitivity.
  $n/K$ counts questions/paired hop conditions, not model calls.
  All score columns are item-macro condition means multiplied by 100;
  $C-D$ is Control minus Drop, with a paired question-bootstrap interval.
  Length bands additionally require sentence-aligned deletion.}
  \label{tab:frames-background-sensitivity}
  \begin{tabular}{@{}llccccc@{}}
    \toprule
    Model & Filter & $n/K$ & Full & Control & Drop & $C-D$ [95\% CI] \\
    \midrule
    DeepSeek V4 Flash & All feasible & 80/204 & 77.50 & 78.09 & 9.81 & 68.3 [60.9, 75.6] \\
      & $\pm25\%$ & 80/191 & 77.50 & 77.74 & 7.92 & 69.8 [62.7, 76.8] \\
      & $\pm10\%$ & 56/78  & 74.40 & 77.38 & 9.33 & 68.1 [57.9, 77.7] \\
      & $\pm5\%$  & 30/38  & 74.44 & 80.56 & 12.78 & 67.8 [53.3, 81.1] \\
    \midrule
    Gemini 3.6 Flash & All feasible & 81/207 & 86.01 & 83.82 & 9.37 & 74.5 [66.9, 81.6] \\
      & $\pm25\%$ & 81/194 & 86.01 & 84.40 & 7.72 & 76.7 [69.2, 83.8] \\
      & $\pm10\%$ & 57/80  & 85.38 & 83.92 & 10.04 & 73.9 [63.5, 83.8] \\
      & $\pm5\%$  & 30/38  & 84.44 & 83.15 & 13.33 & 69.8 [53.9, 84.8] \\
    \midrule
    Qwen 3.7 Max & All feasible & 79/201 & 68.57 & 67.98 & 8.09 & 59.9 [50.8, 68.6] \\
      & $\pm25\%$ & 79/188 & 68.57 & 68.11 & 6.19 & 61.9 [53.2, 70.5] \\
      & $\pm10\%$ & 55/77  & 68.79 & 68.08 & 8.38 & 59.7 [47.8, 70.8] \\
      & $\pm5\%$  & 30/38  & 71.67 & 68.33 & 7.78 & 60.6 [42.8, 77.2] \\
    \bottomrule
  \end{tabular}
\end{table}

All three constructed-side Control--Drop intervals remain above zero in
every band (Table~\ref{tab:frames-background-sensitivity}). Even the 5\% band
retains 38 conditions from 30 questions per model and a 60.6--69.8-point gap.
The tighter subsets change the question population and do not establish exact
matching of passage position, deletion geometry, or semantic relevance.
Together with the common-Full check, the results support increased measured
evidence dependence that cannot be attributed solely to differing answerable
populations or nonspecific text deletion; they do not establish universal
necessity of every evidence unit.

\paragraph{Uncertainty and reproducibility.}
All supplementary intervals use 10{,}000 whole-question bootstrap resamples
and percentile endpoints, with base seed 20260925 and fixed model/cohort
offsets. For BNR, each resampled question retains all hops, trials, both sides,
and both Full-correct definitions; the intersection and Avg@3 estimates are
recomputed within every resample. For background scores, each side and length
cohort is resampled separately, keeping its Full/Control/Drop triple paired.
These intervals reflect question-sampling variability conditional on the
saved scores, not independent uncertainty in model trials, judges, or generated
contexts. No multiplicity correction or equivalence test is claimed.
The standard-library script \texttt{offline\_matched\_analysis.py} records
input and script hashes, per-condition matches, exclusion lists, all original-
and constructed-side sensitivity results, and bootstrap settings in its JSON
output. This is independent of the MuSiQue background-control experiment in
Appendix~\ref{app:matched-controls}.

\section{Human Spot Checks}
\label{app:human-spotchecks}

After adjudication, spot checks accepted 98/100 frozen MuSiQue items and
50/50 LongBench items. Protocol and disagreements are below.

\subsection{Human Spot Check of \realhopmusique{}}
\label{app:musique-spotcheck}

After freezing the 790-item evaluation set, we sampled 100
\realhopmusique{} instances, hop-stratified at random with quotas
54/31/15, matching the 430/244/116 split of 2-, 3-, and 4-hop
items. Annotators received the question, labeled answer, realized
context, gold evidence, and inserted competitor sentences; model outputs
were withheld.

For each item, annotators judged whether the written gold evidence matches
the question and supports the labeled answer, and whether each inserted
competitor relation fails to entail the required gold relation, so those
sentences do not yield that answer. Labeled answers are newly coined names
bound to the constructed entities; retained source text concerns the
original entities and cannot recover these answers.

Two annotators labeled all 100 items. They marked 4 and 5 items as
questionable, and agreed on 95/100: both accepted 93 items, both rejected
2, and disagreed on 5. A third annotator judged the disagreements
correct. After that, two items (2/100) have gold-chain errors from a
pronoun with the wrong antecedent; the competitor sentences were not
judged to give the gold answer.

\subsection{Human Spot Check of \realhoplongbench{}}
\label{app:longbench-spotcheck}

We applied the same protocol to 50 randomly sampled
\realhoplongbench{} instances from the frozen 216-item set. Constructed
answers are new names absent from the unmodified carrier text, so residual
carrier passages cannot recover them.

Two annotators labeled all 50 items. They marked 1 and 0 items as
questionable, and agreed on 49/50: both accepted 49 items and disagreed
on 1. A third annotator judged the disagreement correct. After that, all
50 items were accepted.

The higher acceptance rate is consistent with a difference in
realization. \realhopmusique{} inserts complete sentences while retaining
source text in order, so a later demonstrative can bind to an intervening
source entity; the two residual errors in the MuSiQue sample are of this
kind. \realhoplongbench{} rewrites the local neighborhood of each
insertion, so gold and competitor relations are realized with explicit
antecedents rather than cross-sentence pronouns into unmodified carrier
text.

\section{Evaluation Prompts}
\label{app:prompts}

The scores in the paper use the frozen templates below. The judge sees the
question, reference answers, and model response, and does not see the
document. For \realhopmusique{}, HotpotQA, 2WikiMultiHopQA, and
\realhoplongbench{}, only verdict \texttt{correct} counts as correct.
\texttt{partial} is not treated as correct. FRAMES uses the same judge
template; its recorded \texttt{correct}/\texttt{partial}/\texttt{incorrect}
verdicts map to $1$/$0.5$/$0$. The reported FRAMES Full accuracy and BNR
binarize those recorded scores, counting only $1$ as correct
(Appendix~\ref{app:frames-bnr-trials}).

\paragraph{Multi-hop solver.}
\realhopmusique{}, source MuSiQue, HotpotQA, and 2WikiMultiHopQA use the
official LongBench v1 HotpotQA template. FRAMES uses this template for the
reported Full and Drop conditions.

\begin{lstlisting}[style=prompt]
Answer the question based on the given passages. Only give me the answer and do not output any other words.

The following are given passages.
{context}

Answer the question based on the given passages. Only give me the answer and do not output any other words.

Question: {input}
Answer:
\end{lstlisting}

\paragraph{Long-context open-ended solver.}
\realhoplongbench{} Full, Drop, and random-window trials use:

\begin{lstlisting}[style=prompt]
Please read the following text and answer the question below.


{context}


Question: {question}

Give only the concise answer. Do not provide an explanation or list alternatives.
\end{lstlisting}

Normalized exact matches are accepted without a judge call; remaining
responses use the short-answer judge below.

\paragraph{Long-context multiple-choice solver.}
Matched source questions and the \realhoplongbench{} MCQ evaluation use the
same official LongBench v2 0-shot multiple-choice template. The constructed
evaluation keeps a frozen A--D option order, with the gold terminal and three
planned competitor terminals as the four choices.

\begin{lstlisting}[style=prompt]
Please read the following text and answer the question below.

$DOC$

What is the correct answer to this question: $Q$
Choices:
(A) $C_A$
(B) $C_B$
(C) $C_C$
(D) $C_D$

Format your response as follows: "The correct answer is (insert answer here)".
\end{lstlisting}

\paragraph{Short-answer judge.}
The short-answer judge uses the following system message:

\begin{lstlisting}[style=prompt]
You are a calibrated grader for short-answer multi-hop QA. Grade the response relative to the QUESTION, not as a standalone sentence. The reference answer may be more verbose than the minimum answer actually requested by the question.
Reply with a SINGLE LINE of JSON in this exact format and NOTHING ELSE:
{"verdict": "<correct|partial|incorrect>", "reason": "<one short sentence>"}
Grading rules:
  - correct: the response fully fills the answer slot requested by the question and is semantically equivalent to a reference. Allow paraphrases, explicit aliases, harmless generic prefixes or suffixes, and omission of participants or roles already unambiguously fixed by the question.
  - A pronoun or short predicate can be complete when the question itself uniquely supplies its participant. For example, for 'How were X expelled?', 'defeated' can be complete; do not require repeating X or the actor unless the question asks who.
  - Preserve answer-defining distinctions not supplied by the question. A wrong entity, location, date, direction, number, role, scope, or predicate is incorrect.
  - Do not accept a merely related detail in place of the requested slot. For example, if asked what role or category someone won for, a work title alone is partial or incorrect.
  - Extra information is allowed unless it contradicts or changes the answer.
  - partial: the response gives a genuinely correct part of the requested answer but omits a detail needed to fill the requested slot.
  - incorrect: the response is wrong, unsupported, contradictory, refuses to answer, or contains only topical overlap.
  - Judge only from the ordinary semantics of the question, references, and response. Do not assume hidden source context or an unstated alias.
Do not output anything other than the JSON line.
\end{lstlisting}

User message:

\begin{lstlisting}[style=prompt]
Question:
{question}

Reference answer(s) (any one suffices):
{references}

Model response:
{response}

Now grade the model response.
\end{lstlisting}

\end{document}